\documentclass[runningheads]{llncs}
\usepackage{graphicx}
\usepackage{amsmath,amssymb} 
\usepackage{color}
\usepackage[width=122mm,left=12mm,paperwidth=146mm,height=193mm,top=12mm,paperheight=217mm]{geometry}

\def\eg{\emph{e.g.}}

\def\ie{\emph{i.e.}}

\def\vs{\emph{vs. }}

\usepackage{subfig}

\begin{document}
\pagestyle{headings}
\mainmatter

\title{Is Faster R-CNN Doing Well for \\ Pedestrian Detection?} 

\titlerunning{~}

\authorrunning{~}




\author{Liliang~Zhang$^{1}$  \quad Liang~Lin$^{1}$\thanks{The corresponding author is Liang Lin.}\quad Xiaodan~Liang$^{1}$  \quad Kaiming~He$^{2}$}

\institute{$^{1}$School of Data and Computer Science, Sun Yat-sen University \\
\email{zhangll.level0@gmail.com; linliang@ieee.org; xdliang328@gmail.com} \\
$^{2}$Microsoft Research \\
\email{kahe@microsoft.com} 
}

\maketitle

\begin{abstract}
Detecting pedestrian has been arguably addressed as a special topic beyond general object detection. Although recent deep learning object detectors such as Fast/Faster R-CNN \cite{girshickICCV15fastrcnn,ren2015faster} have shown excellent performance for general object detection, they have limited success for detecting pedestrian, and previous leading pedestrian detectors were in general hybrid methods combining hand-crafted and deep convolutional features. In this paper, we investigate issues involving Faster R-CNN \cite{ren2015faster} for pedestrian detection. We discover that the Region Proposal Network (RPN) in Faster R-CNN indeed performs well as a stand-alone pedestrian detector, but surprisingly, the downstream classifier degrades the results. We argue that two reasons account for the unsatisfactory accuracy: (i) insufficient resolution of feature maps for handling small instances, and (ii) lack of any bootstrapping strategy for mining hard negative examples. Driven by these observations, we propose a very simple but effective baseline for pedestrian detection, using an RPN followed by boosted forests on shared, high-resolution convolutional feature maps. We comprehensively evaluate this method on several benchmarks (Caltech, INRIA, ETH, and KITTI), presenting competitive accuracy and good speed. Code will be made publicly available.

\end{abstract}

\section{Introduction}

Pedestrian detection, as a key component of real-world applications such as automatic driving and intelligent surveillance, has attracted special attention beyond general object detection. Despite the prevalent success of deeply learned features in computer vision, current leading pedestrian detectors (\eg, \cite{hosang2015taking,tian2015pedestrian,tian2015deep,cai2015learning}) are in general \emph{hybrid} methods that combines traditional, hand-crafted features \cite{dollar2009integral,dollar2014fast} and deep convolutional features \cite{krizhevsky2012imagenet,simonyan2014very}. For example, in \cite{hosang2015taking} a stand-alone pedestrian detector \cite{benenson2014ten} (that uses Squares Channel Features) is adopted as a highly selective proposer ($<$3 regions per image), followed by R-CNN \cite{girshick2014rich} for classification. Hand-crafted features appear to be of critical importance for state-of-the-art pedestrian detection.

On the other hand, Faster R-CNN \cite{ren2015faster} is a particularly successful method for general object detection. It consists of two components: a fully convolutional Region Proposal Network (RPN) for proposing candidate regions, followed by a downstream Fast R-CNN \cite{girshickICCV15fastrcnn} classifier. The Faster R-CNN system is thus a purely CNN-based method without using hand-crafted features (\eg, Selective Search \cite{uijlings2013selective} that is based on low-level features). Despite its leading accuracy on several multi-category benchmarks, Faster R-CNN has not presented competitive results on popular pedestrian detection datasets (\eg, the Caltech set \cite{dollar2012pedestrian}).

In this paper, we investigate the issues involving Faster R-CNN as a pedestrian detector. Interestingly, we find that an RPN specially tailored for pedestrian detection achieves competitive results as a stand-alone pedestrian detector. But surprisingly, the accuracy is degraded after feeding these proposals into the Fast R-CNN classifier. We argue that such unsatisfactory performance is attributed to two reasons as follows.

\begin{figure}[t] \centering
\subfloat[Small positive instances]{ \centering
\includegraphics[width=0.45\textwidth]{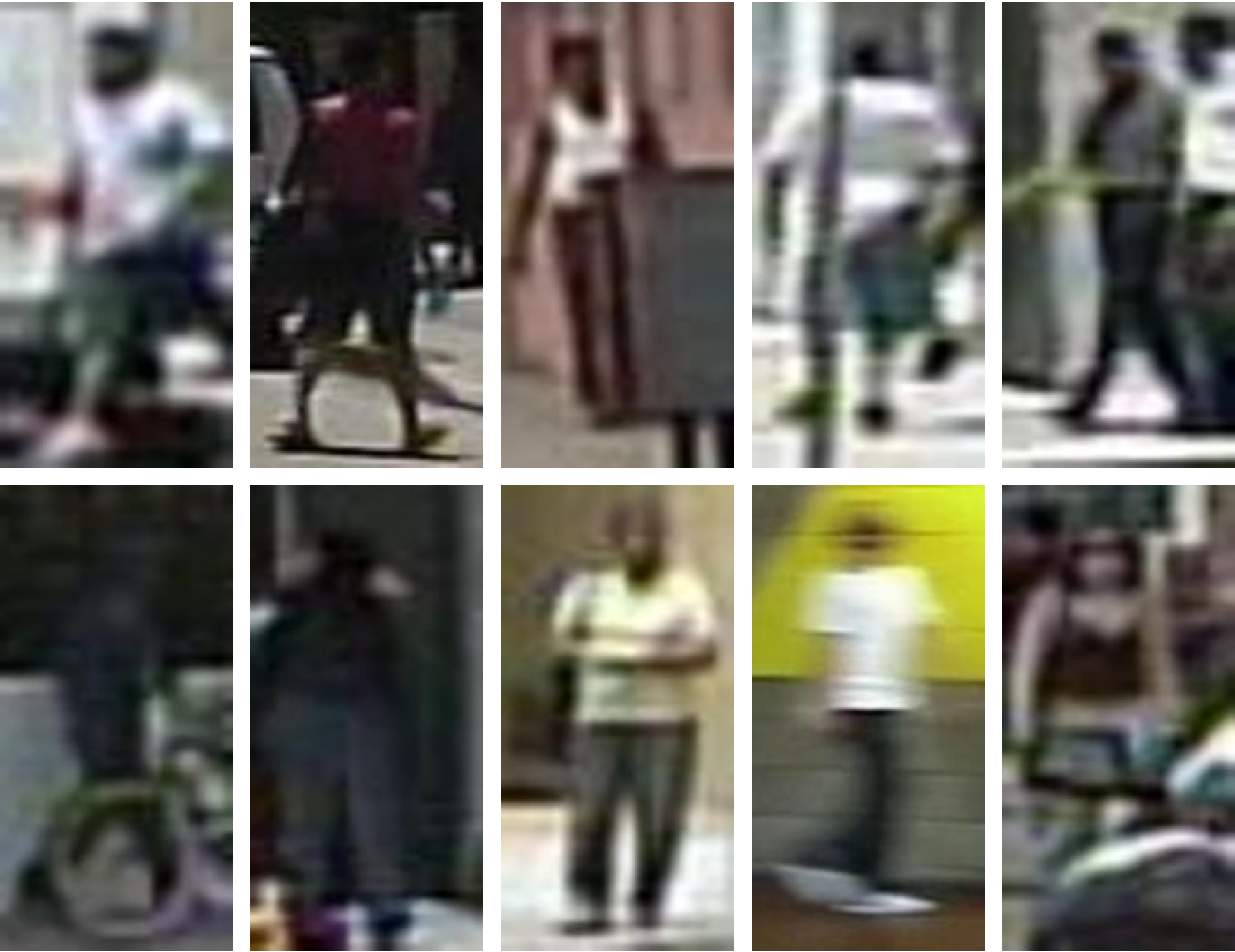}
\label{fig:tp}
}\subfloat[Hard negatives]{ \centering
\includegraphics[width=0.45\textwidth]{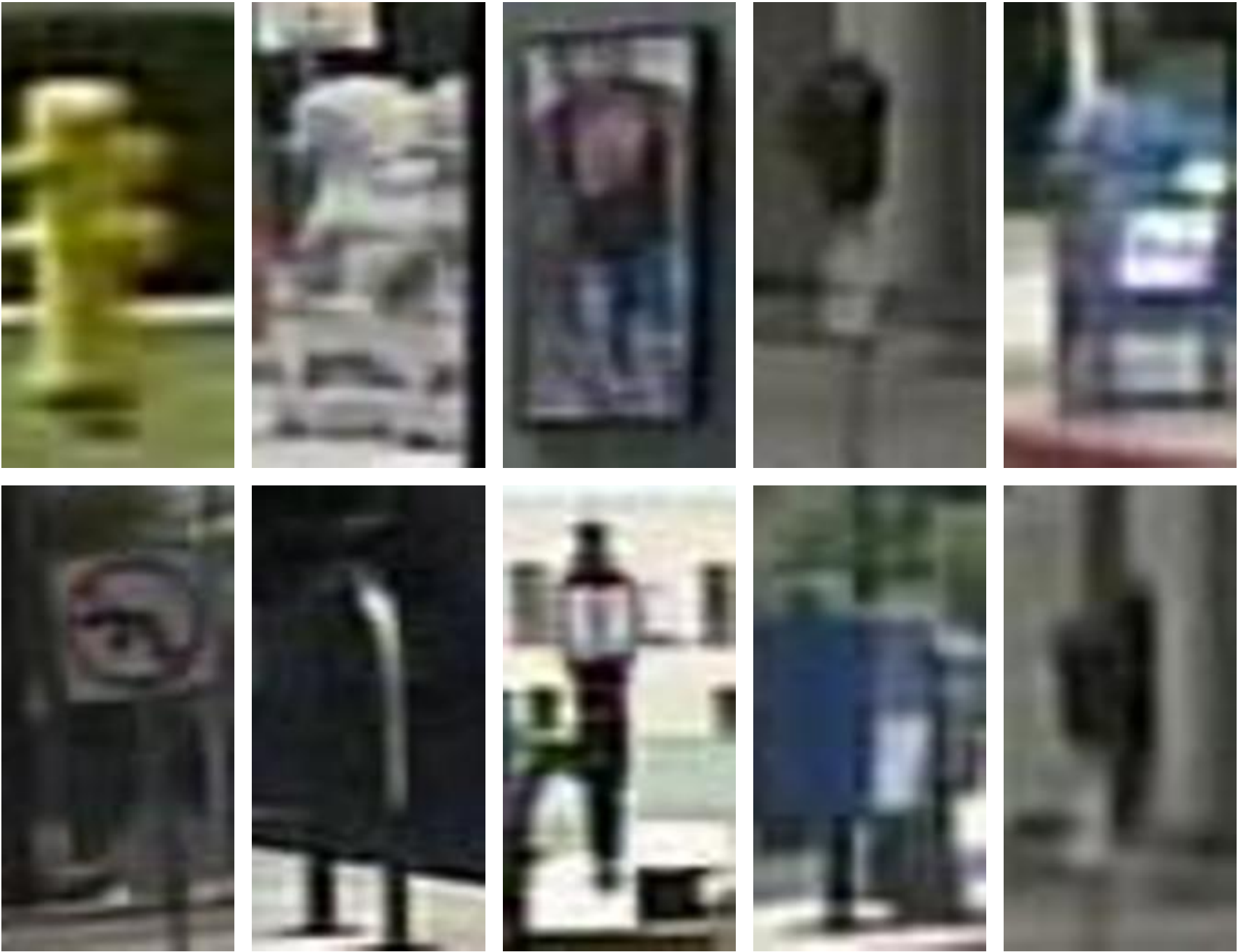}
\label{fig:fp-bg}
}
\caption{Two challenges for Fast/Faster R-CNN in pedestrian detection. (a) Small objects that may fail RoI pooling on low-resolution feature maps. (b) Hard negative examples that receive no careful attention in Fast/Faster R-CNN.}
\label{fig:hard-neg-example}
\end{figure}

First, the convolutional feature maps of the Fast R-CNN classifier are of low solution for detecting small objects. Typical scenarios of pedestrian detection, such as automatic driving and intellegent surveillance, generally present pedestrian instances of small sizes (\eg, 28$\times$70 for Caltech \cite{dollar2012pedestrian}). On small objects (Fig.~\ref{fig:hard-neg-example}(a)), the Region-of-Interest (RoI) pooling layer \cite{he14ECCV,girshickICCV15fastrcnn} performed on a low-resolution feature map (usually with a stride of 16 pixels) can lead to ``plain'' features caused by collapsing bins. These features are not discriminative on small regions, and thus degrade the downstream classifier. We note that this is in contrast to hand-crafted features that have finer resolutions.
We address this problem by pooling features from shallower but higher-resolution layers, and by the hole algorithm (namely, ``{\`a} trous'' \cite{chen2014semantic} or filter rarefaction \cite{long2015fully}) that increases feature map size.

Second, in pedestrian detection the false predictions are dominantly caused by confusions of hard \emph{background} instances (Fig.~\ref{fig:hard-neg-example}(b)). This is in contrast to general object detection where a main source of confusion is from \emph{multiple categories}. To address hard negative examples, we adopt cascaded Boosted Forest (BF) \cite{friedman2000additive,appel2013quickly}, which performs effective hard negative mining (bootstrapping) and sample re-weighting, to classify the RPN proposals. Unlike previous methods that use hand-crafted features to train the forest, in our method the BF \emph{reuses} the deep convolutional features of RPN. This strategy not only reduces the computational cost of the classifier by sharing features, but also exploits the deeply learned features.

As such, we present a surprisingly simple but effective baseline for pedestrian detection based on RPN and BF. Our method overcomes two limitations of Faster R-CNN for pedestrian detection and gets rid of traditional hand-crafted features. We present compelling results on several benchmarks, including Caltech \cite{dollar2012pedestrian}, INRIA \cite{dalal2005histograms}, ETH \cite{ess2007depth}, and KITTI \cite{geiger2012kitti}. Remarkably, our method has substantially better localization accuracy and shows a relative improvement of 40\% on the Caltech dataset under an Intersection-over-Union (IoU) threshold of 0.7 for evaluation.
Meanwhile, our method has a test-time speed of 0.5 second per image, which is competitive with previous leading methods.

In addition, our paper reveals that traditional pedestrian detectors have been inherited in recent methods at least for two reasons. First, the higher resolution of hand-crafted features (such as \cite{dollar2009integral,dollar2014fast}) and their pyramids is good for detecting small objects. Second, effective bootstrapping is performed for mining hard negative examples. These key factors, however, when appropriately handled in a deep learning system, lead to excellent results.

\begin{figure*}[t]
\centering
\includegraphics[width=0.90\textwidth]{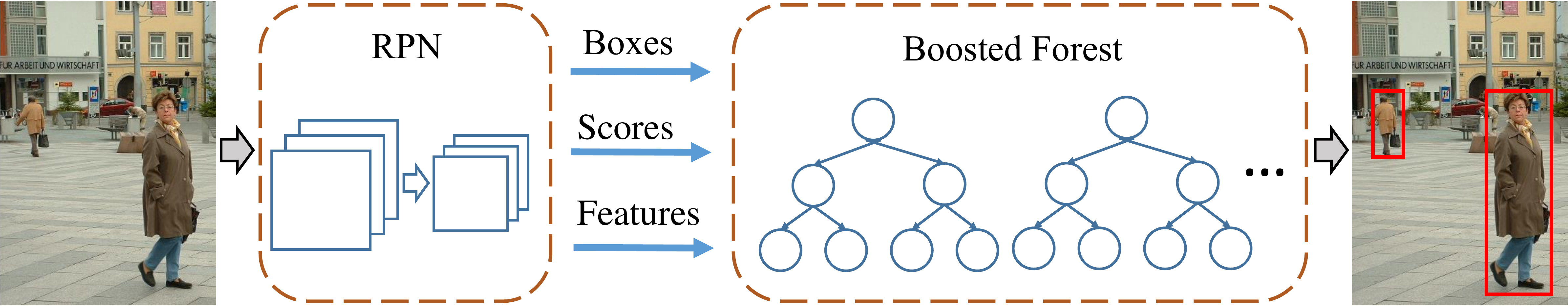}
\caption{Our pipeline. RPN is used to compute candidate bounding boxes, scores, and convolutional feature maps. The candidate boxes are fed into cascaded Boosted Forests (BF) for classification, using the features pooled from the convolutional feature maps computed by RPN.}
\label{fig:overall-pipeline}
\end{figure*}

\section{Related Work}

The Integrate Channel Features (ICF) detector \cite{dollar2009integral}, which extends the Viola-Jones framework \cite{viola2004robust}, is among the most popular pedestrian detectors without using deep learning features.
The ICF detector involves channel feature pyramids and boosted classifiers. 
The feature representations of ICF have been improved in several ways, including ACF \cite{dollar2014fast}, LDCF \cite{nam2014local}, SCF \cite{benenson2014ten}, and many others, but the boosting algorithm remains a key building block for pedestrian detection.

Driven by the success of (``slow'') R-CNN \cite{girshick2014rich} for general object detection, a recent series of methods \cite{benenson2014ten,tian2015pedestrian,tian2015deep} adopt a two-stage pipeline for pedestrian detection. In \cite{hosang2015taking}, the SCF pedestrian detector \cite{benenson2014ten} is used to propose regions, followed by an R-CNN for classification; TA-CNN \cite{tian2015pedestrian} employs the ACF detector \cite{dollar2014fast} to generate proposals, and trains an R-CNN-style network to jointly optimize pedestrian detection with semantic tasks; the DeepParts method \cite{tian2015deep} applies the LDCF detector \cite{nam2014local} to generate proposals and learns a set of complementary parts by neural networks. We note that these proposers are stand-alone pedestrian detectors consisting of hand-crafted features and boosted classifiers.

Unlike the above R-CNN-based methods, the CompACT method \cite{cai2015learning} learns boosted classifiers on top of hybrid hand-crafted and deep convolutional features. Most closely related to our work, the CCF detector \cite{yang2015convolutional} is boosted classifiers on pyramids of deep convolutional features, but uses no region proposals. Our method has no pyramid, and is much faster and more accurate than \cite{yang2015convolutional}.

\section{Approach}

Our approach consists of two components (illustrated in Fig.~\ref{fig:overall-pipeline}): an RPN that generates candidate boxes as well as convolutional feature maps, and a Boosted Forest that classifies these proposals using these convolutional features.

\subsection{Region Proposal Network for Pedestrian Detection}

The RPN in Faster R-CNN \cite{ren2015faster} was developed as a class-agnostic detector (proposer) in the scenario of multi-category object detection. For single-category detection, RPN is naturally a detector for the only category concerned. We specially tailor the RPN for pedestrian detection, as introduced in the following. 

We adopt anchors (reference boxes) \cite{ren2015faster} of a single aspect ratio of 0.41 (width to height). This is the average aspect ratio of pedestrians as indicated in \cite{dollar2012pedestrian}. This is unlike the original RPN \cite{ren2015faster} that has anchors of multiple aspect ratios. Anchors of inappropriate aspect ratios are associated with few examples, so are noisy and harmful for detection accuracy.
In addition, we use anchors of 9 different scales, starting from 40 pixels height with a scaling stride of 1.3$\times$. This spans a wider range of scales than \cite{ren2015faster}. The usage of multi-scale anchors waives the requirement of using feature pyramids to detect multi-scale objects.

Following \cite{ren2015faster}, we adopt the VGG-16 net \cite{simonyan2014very} pre-trained on the ImageNet dataset \cite{ILSVRC15} as the backbone network. The RPN is built on top of the Conv5\_3 layer, which is followed by an intermediate 3$\times$3 convolutional layer and two sibling 1$\times$1 convolutional layers for classification and bounding box regression (more details in \cite{ren2015faster}). In this way, RPN regresses boxes with a stride of 16 pixels (Conv5\_3). The classification layer provides confidence scores of the predicted boxes, which can be used as the initial scores of the Boosted Forest cascade that follows.

It is noteworthy that although we will use the ``{\`a} trous'' \cite{chen2014semantic} trick in the following section to increase resolution and reduce stride, we keep using the same RPN with a stride of 16 pixels. The {\`a} trous trick is only exploited when extracting features (as introduced next), but not for fine-tuning.

\subsection{Feature Extraction}
\label{sec:feature}

With the proposals generated by RPN, we adopt RoI pooling \cite{girshickICCV15fastrcnn} to extract fixed-length features from regions. These features will be used to train BF as introduced in the next section. Unlike Faster R-CNN which requires to feed these features into the \emph{original fully-connected} (fc) layers and thus limits their dimensions, the BF classifier imposes no constraint on the dimensions of features. For example, we can extract features from RoIs on Conv3\_3 (of a stride = 4 pixels) and Conv4\_3 (of a stride = 8 pixels). We pool the features into a fixed resolution of 7$\times$7. These features from different layers are simply concatenated without normalization, thanks to the flexibility of the BF classifier; on the contrast, feature normalization needs to be carefully addressed \cite{liu2015parsenet} for deep classifiers when concatenating features.

Remarkably, as there is no constraint imposed to feature dimensions, it is flexible for us to use features of increased resolution. In particular, given the fine-tuned layers from RPN (stride = 4 on Conv3,  8 on Conv4, and 16 on Conv5), we can use the {\`a} trous trick \cite{chen2014semantic} to compute convolutional feature maps of higher resolution. For example, we can set the stride of Pool3 as 1 and dilate all Conv4 filters by 2, which reduces the stride of Conv4 from 8 to 4. Unlike previous methods \cite{long2015fully,chen2014semantic} that fine-tune the dilated filters, in our method we only use them for feature extraction, without fine-tuning a new RPN. 

Though we adopt the same RoI resolution (7$\times$7) as Faster R-CNN \cite{ren2015faster}, these RoIs are on higher-resolution feature maps (\eg, Conv3\_3, Conv4\_3, or Conv4\_3 {\`a}~trous) than Fast R-CNN (Conv5\_3). If an RoI's input resolution is smaller than output (\ie, $< 7\times7$), the pooling bins collapse and the features become ``flat'' and not discriminative. This problem is alleviated in our method, as it is not constrained to use features of Conv5\_3 in our downstream classifier.

\subsection{Boosted Forest}

The RPN has generated the region proposals, confidence scores, and features, all of which are used to train a cascaded Boosted Forest classifier. We adopt the RealBoost algorithm \cite{friedman2000additive}, and mainly follow the hyper-parameters in \cite{cai2015learning}.
Formally, we bootstrap the training by 6 times, and the forest in each stage has $\{64, 128, 256, 512, 1024, 1536\}$ trees. Initially, the training set consists of all positive examples ($\sim$50k on the Caltech set) and the same number of randomly sampled negative examples from the proposals. After each stage, additional hard negative examples (whose number is 10\% of the positives, $\sim$5k on Caltech) are mined and added into the training set. Finally, a forest of 2048 trees is trained after all bootstrapping stages. This final forest classifier is used for inference. Our implementation is based on \cite{PMT}.

We note that it is not necessary to handle the initial proposals equally, because our proposals have initial confidence scores computed by RPN. In other words, the RPN can be considered as the stage-0 classifier $f_0$, and we set $f_{0} = \frac{1}{2} \log \frac {s} {1 - s}$ following the RealBoost form where $s$ is the score of a proposal region ($f_0$ is a constant in standard boosting). The other stages are as in standard RealBoost.

\subsection{Implementation Details}

We adopt single-scale training and testing as in \cite{he14ECCV,girshickICCV15fastrcnn,ren2015faster}, without using feature pyramids.
An image is resized such that its shorter edge has $N$ pixels ($N$=720 pixels on Caltech, 600 on INRIA, 810 on ETH, and 500 on KITTI).
For RPN training, an anchor is considered as a positive example if it has an Intersection-over-Union (IoU) ratio greater than $0.5$ with one ground truth box, and otherwise negative. We adopt the image-centric training scheme \cite{girshickICCV15fastrcnn,ren2015faster}, and each mini-batch consists of 1 image and 120 randomly sampled anchors for computing the loss.  The ratio of positive and negative samples is 1:5 in a mini-batch. Other hyper-parameters of RPN are as in \cite{ren2015faster}, and we adopt the publicly available code of \cite{ren2015faster} to fine-tune the RPN. We note that in \cite{ren2015faster} the cross-boundary anchors are ignored during fine-tuning, whereas in our implementation we preserve the cross-boundary negative anchors during fine-tuning, which empirically improves accuracy on these datasets.

With the fine-tuned RPN, we adopt non-maximum suppression (NMS) with a threshold of 0.7 to filter the proposal regions. Then the proposal regions are ranked by their scores. 
For BF training, we construct the training set by selecting the top-ranked 1000 proposals (and ground truths) of each image. The tree depth is set as 5 for the Caltech and KITTI set, and 2 for the INRIA and ETH set, which are empirically determined according to the different sizes of the data sets. At test time, we only use the top-ranked 100 proposals in an image, which are classified by the BF.

\section{Experiments and Analysis}

\subsection{Datasets}

We comprehensively evaluate on 4 benchmarks: Caltech \cite{dollar2012pedestrian}, INRIA \cite{dalal2005histograms}, ETH \cite{ess2007depth} and KITTI \cite{geiger2012kitti}. By default an IoU threshold of 0.5 is used for determining True Positives in these datasets.

On Caltech \cite{dollar2012pedestrian}, the training data is augmented by 10 folds (42782 images) following \cite{hosang2015taking}. 4024 images in the standard test set are used for evaluation on the original annotations under the ``reasonable''  setting (pedestrians that are at least 50 pixels tall and at least 65\% visible) \cite{dollar2012pedestrian}. The evaluation metric is log-average Miss Rate on False Positive Per Image (FPPI) in $[10^{-2}, 10^{0}]$ (denoted as MR$_{-2}$ following \cite{Zhang2016}, or in short MR). In addition, we also test our model on the new annotations provided by \cite{Zhang2016}, which correct the errors in the original annotations. This set is denoted as ``Caltech-New''. The evaluation metrics in Caltech-New are MR$_{-2}$ and MR$_{-4}$, corresponding to the log-average Miss Rate on FPPI ranges of $[10^{-2}, 10^{0}]$ and $[10^{-4}, 10^{0}]$, following \cite{Zhang2016}.

The INRIA \cite{dalal2005histograms} and ETH \cite{ess2007depth} datasets are often used for verifying the generalization capability of the models. Following the settings in  \cite{paisitkriangkrai2014strengthening}, our model is trained on the 614 positive and 1218 negative images in the INRIA training set. The models are evaluated on the 288 testing images in INRIA and 1804 images in ETH, evaluated by MR$_{-2}$.

The KITTI dataset \cite{geiger2012kitti} consists of images with stereo data available. We perform training on the 7481 images of the left camera, and evaluate on the standard 7518 test images. 
KITTI evaluates the PASCAL-style mean Average Precision (mAP) under three difficulty levels: ``Easy'', ``Moderate'', and ``Hard''\footnote{http://www.cvlibs.net/datasets/kitti/eval\_object.php}.

\subsection{Ablation Experiments}

In this subsection, we conduct ablation experiments on the Caltech dataset.

\subsubsection{Is RPN good for pedestrian detection?}\hfill

\begin{figure}[t] \centering
\subfloat[1 proposal]{ \centering
\includegraphics[width=0.33\textwidth]{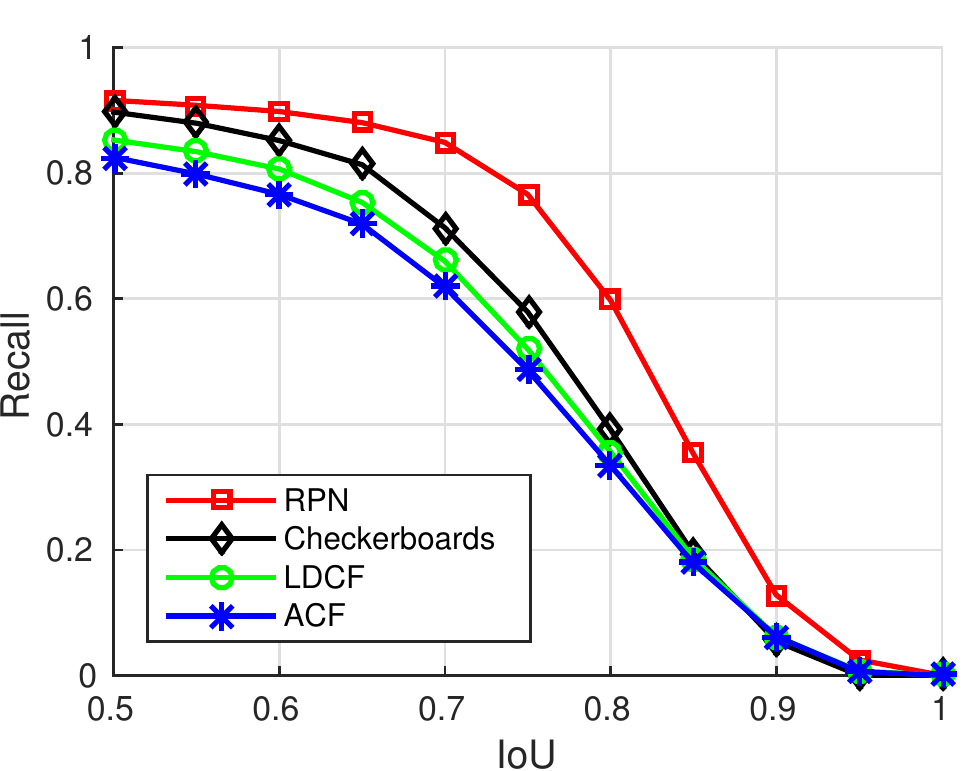}
\label{fig:proposal1}
}\subfloat[4 proposals]{ \centering
\includegraphics[width=0.33\textwidth]{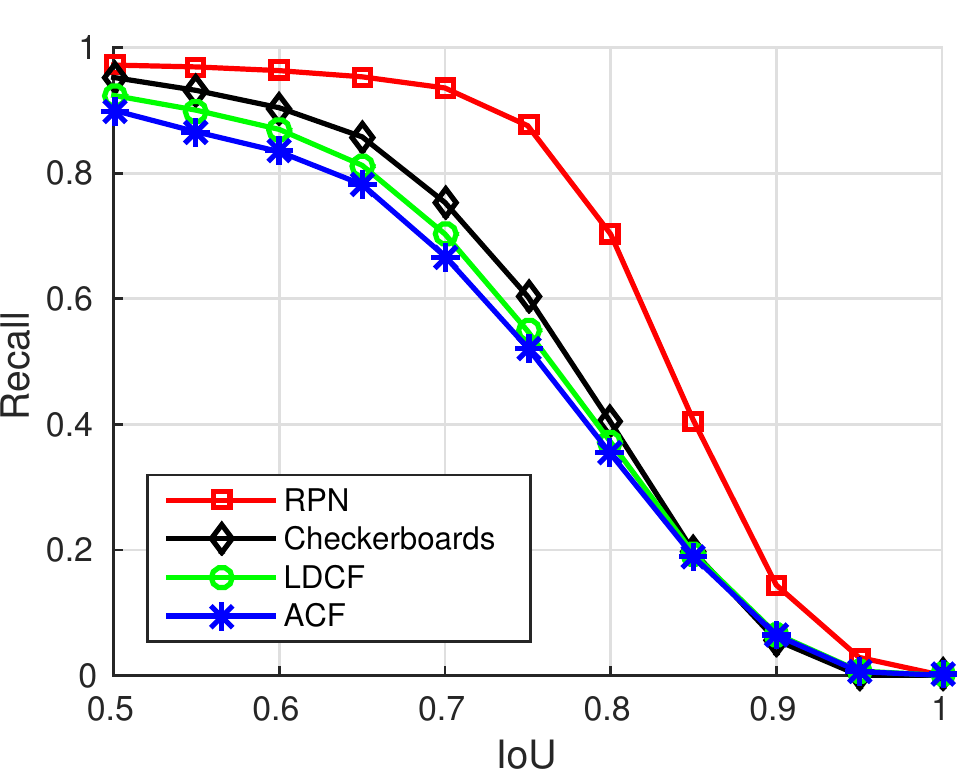}
\label{fig:proposal4}
}\subfloat[100 proposals]{ \centering
\includegraphics[width=0.33\textwidth]{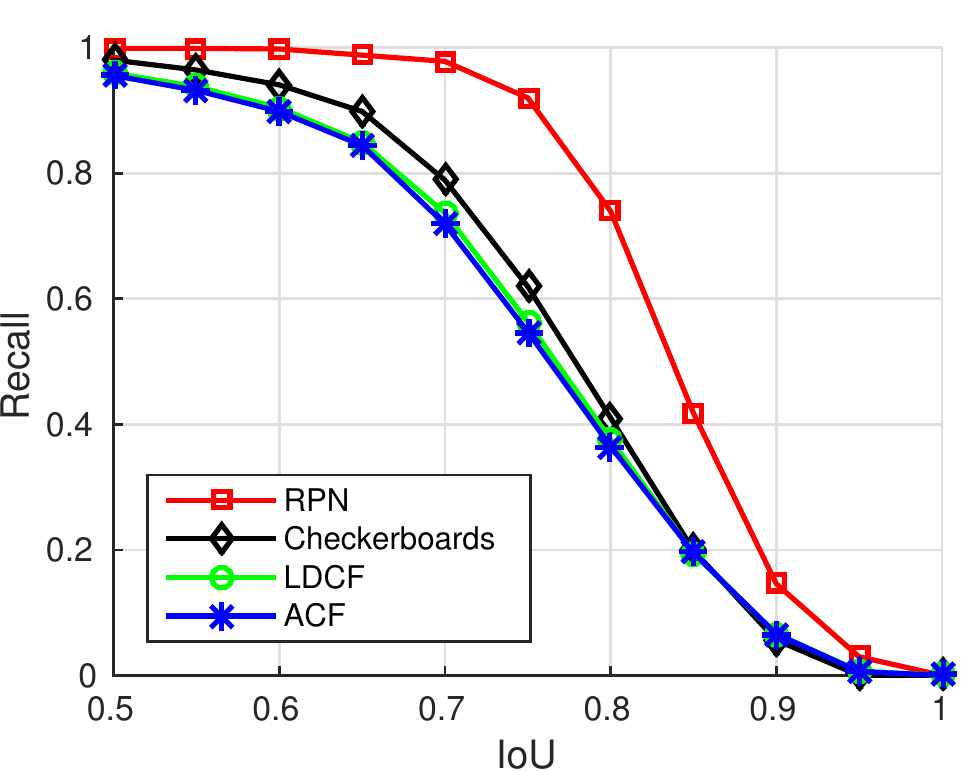}
\label{fig:proposal100}
}
\caption{Comparison of RPN and three existing methods in terms of proposal quality (recall \vs IoU) on the Caltech set, with on average 1, 4 or 100 proposals per image are evaluated.}
\label{fig:proposal}
\end{figure}

In Fig.~\ref{fig:proposal} we investigate RPN in terms of \emph{proposal quality}, evaluated by the recall rates under different IoU thresholds. We evaluate on average 1, 4, or 100 proposals per image\footnote{To be precise, ``on average $k$ proposals per image'' means that for a dataset with $M$ images, the top-ranked $kM$ proposals are taken to evaluate the recall.}. Fig.~\ref{fig:proposal} shows that in general RPN performs better than three leading methods that are based on traditional features: SCF \cite{benenson2014ten}, LDCF \cite{nam2014local} and Checkerboards \cite{zhang2015filtered}.
With 100 proposals per image, our RPN achieves $>$95\% recall at an IoU of 0.7.

More importantly, RPN as a \emph{stand-alone pedestrian detector} achieves an MR of \textbf{14.9\%} (Table~\ref{table-classifier}). \emph{This result is competitive and is better than all but two state-of-the-art competitors on the Caltech dataset} (Fig.~\ref{fig:caltech}). We note that unlike RoI pooling that may suffer from small regions, RPN is essentially based on fixed-size sliding windows (in a fully convolutional fashion) and thus avoids collapsing bins. RPN predicts small objects by using small anchors.

\renewcommand\arraystretch{1.2}
\setlength{\tabcolsep}{6pt}
\begin{table}[h]
\begin{center}
\begin{tabular}{l|c|c}
\hline
method & RoI features & MR (\%) \\
\hline\hline
RPN stand-alone & - & 14.9 \\
\hline
RPN + R-CNN & raw pixels & 13.1 \\
\hline
RPN + Fast R-CNN & Conv5\_3  & 20.2 \\
RPN + Fast R-CNN & Conv5\_3, \`a trous  & 16.2 \\
\hline
RPN + BF & Conv5\_3  & 18.2  \\
RPN + BF & Conv4\_3  & \textbf{12.6}  \\
RPN + BF & Conv5\_3, \`a trous  & 13.7  \\
\hline
\end{tabular}
\end{center}
\caption{Comparisons of different classifiers and features on the Caltech set. All methods are based on VGG-16 (including R-CNN). The same set of RPN proposals are used for all entries. 
}
\label{table-classifier}
\begin{center}
\begin{tabular}{l|c|c}
\hline
RoI features & time/img & MR (\%) \\
\hline\hline
Conv2\_2  & 0.37s &15.9  \\
Conv3\_3  & 0.37s & \textbf{12.4}  \\
Conv4\_3  & 0.37s &12.6  \\
Conv5\_3  & 0.37s &18.2  \\
\hline
Conv3\_3, Conv4\_3 & 0.37s & \textbf{11.5}  \\
Conv3\_3, Conv4\_3, Conv5\_3  & 0.37s & 11.9  \\
\hline
Conv3\_3, (Conv4\_3, \`a trous) & 0.51s & \textbf{9.6}  \\
\hline
\end{tabular}
\end{center}
\caption{Comparisons of different features in our RPN+BF method on the Caltech set. All entries are based on VGG-16 and the same set of RPN proposals.}
\label{table-feature}
\end{table}

\subsubsection{How important is feature resolution?}\hfill

We first report the accuracy of (``slow'') R-CNN \cite{girshick2014rich}. For fair comparisons, we fine-tune R-CNN using the VGG-16 network, and the proposals are from the same RPN as above. This method has an MR of 13.1\% (Table~\ref{table-classifier}), better than its proposals (stand-alone RPN, 14.9\%). R-CNN crops raw pixels from images and warps to a fixed size (224$\times$224), so suffers less from small objects.
This result suggests that if reliable features (\eg, from a fine resolution of 224$\times$224) can be extracted, the downstream classifier is able to improve the accuracy. 

Surprisingly, training a Fast R-CNN classifier on the same set of RPN proposals actually \textit{degrades} the results: the MR is considerably increased to 20.2\% (\vs RPN's 14.9\%, Table~\ref{table-classifier}). Even though R-CNN performs well on this task, Fast R-CNN presents a much worse result.

This problem is partially because of the low-resolution features. To show this, we train a Fast R-CNN (on the same set of RPN proposals as above) with the \`a trous trick adopted on Conv5, reducing the stride from 16 pixels to 8. The problem is alleviated (16.2\%, Table~\ref{table-classifier}), demonstrating that higher resolution can be helpful. Yet, this result still lags far behind the stand-alone RPN or R-CNN (Table~\ref{table-classifier}).

The effects of low-resolution features are also observed in our Boosted Forest classifiers. BF using Conv5\_3 features has an MR of 18.2\% (Table~\ref{table-classifier}), lower than the stand-alone RPN. Using the \`a trous trick on Conv5 when extracting features (Sec.~\ref{sec:feature}), BF has a much better MR of 13.7\%.

But the BF classifier is more flexible and is able to take advantage of features of various resolutions. Table~\ref{table-feature} shows the results of using different features in our method.
Conv3\_3 or Conv4\_3 alone yields good results (12.4\% and 12.6\%), showing the effects of higher resolution features. Conv2\_2 starts to show degradation (15.9\%), which can be explained by the weaker representation of the shallower layers. 
BF on the concatenation of Conv3\_3 and Conv4\_3 features reduces the MR to 11.5\%. The combination of features in this way is nearly cost-free. Moreover, unlike previous usage of skip connections \cite{liu2015parsenet}, it is not necessary to normalize features in a decision forest classifier. 

Finally, combining Conv3\_3 with the \`a trous version of Conv4\_3, we achieve the best result of \textbf{9.6\%} MR. We note that this is at the cost of extra computation (Table~\ref{table-feature}), because it requires to re-compute the Conv4 features maps with the \`a trous trick. Nevertheless, the speed of our method is still competitive (Table~\ref{table-time}).

\subsubsection{How important is bootstrapping?}\hfill

\begin{table}[t]
\begin{center}
\begin{tabular}{l|c|c|c}
\hline
method & RoI features & bootstrapped?& MR (\%) \\
\hline\hline
RPN + Fast R-CNN & Conv5\_3, \`a trous  & & 16.2 \\
RPN + Fast R-CNN & Conv5\_3, \`a trous  & \checkmark & 14.3 \\
\hline
RPN + BF & Conv5\_3, \`a trous  &  \checkmark &  13.7  \\
\hline
\end{tabular}
\end{center}
\caption{Comparisons of with/without bootstrapping on the Caltech set.}
\label{table-bootstrap}
\end{table}

\begin{figure}[t] \centering
\includegraphics[width=1\textwidth]{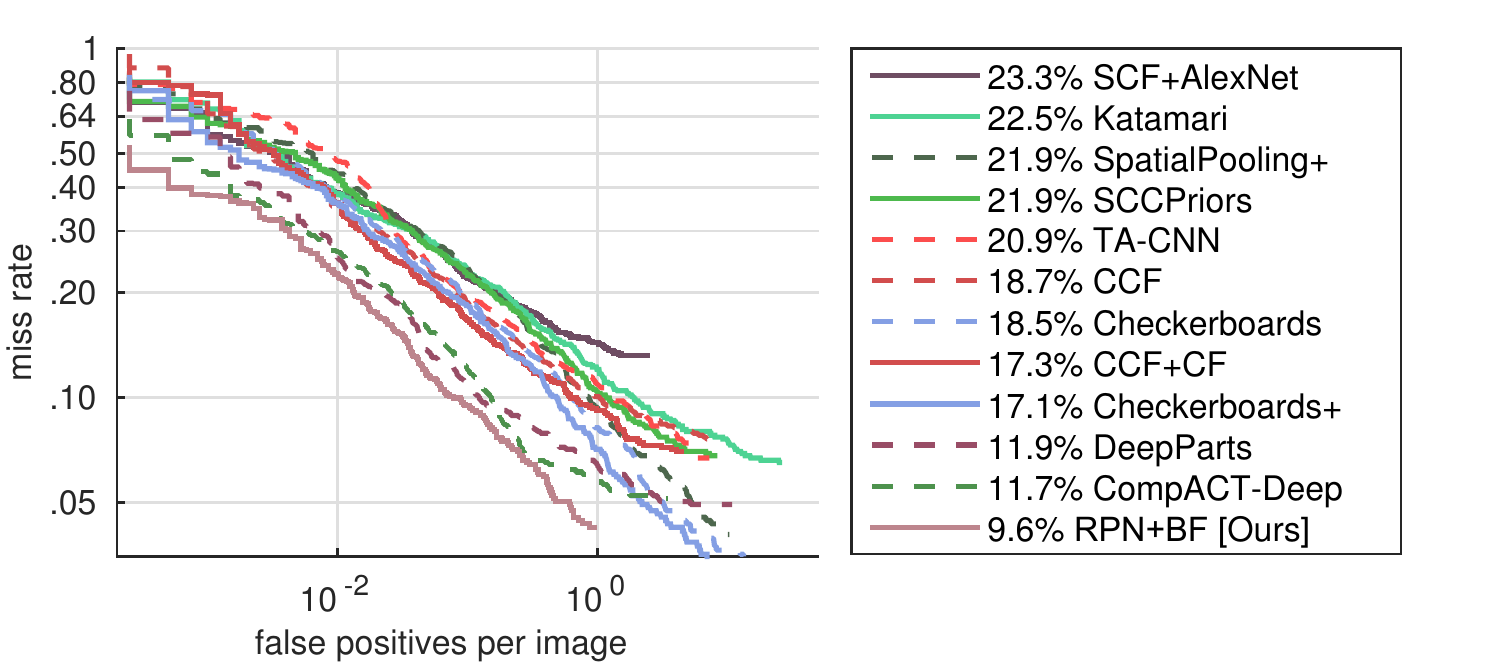}
\caption{Comparisons on the \textbf{Caltech} set (legends indicate MR).}
\label{fig:caltech}
\includegraphics[width=1\textwidth]{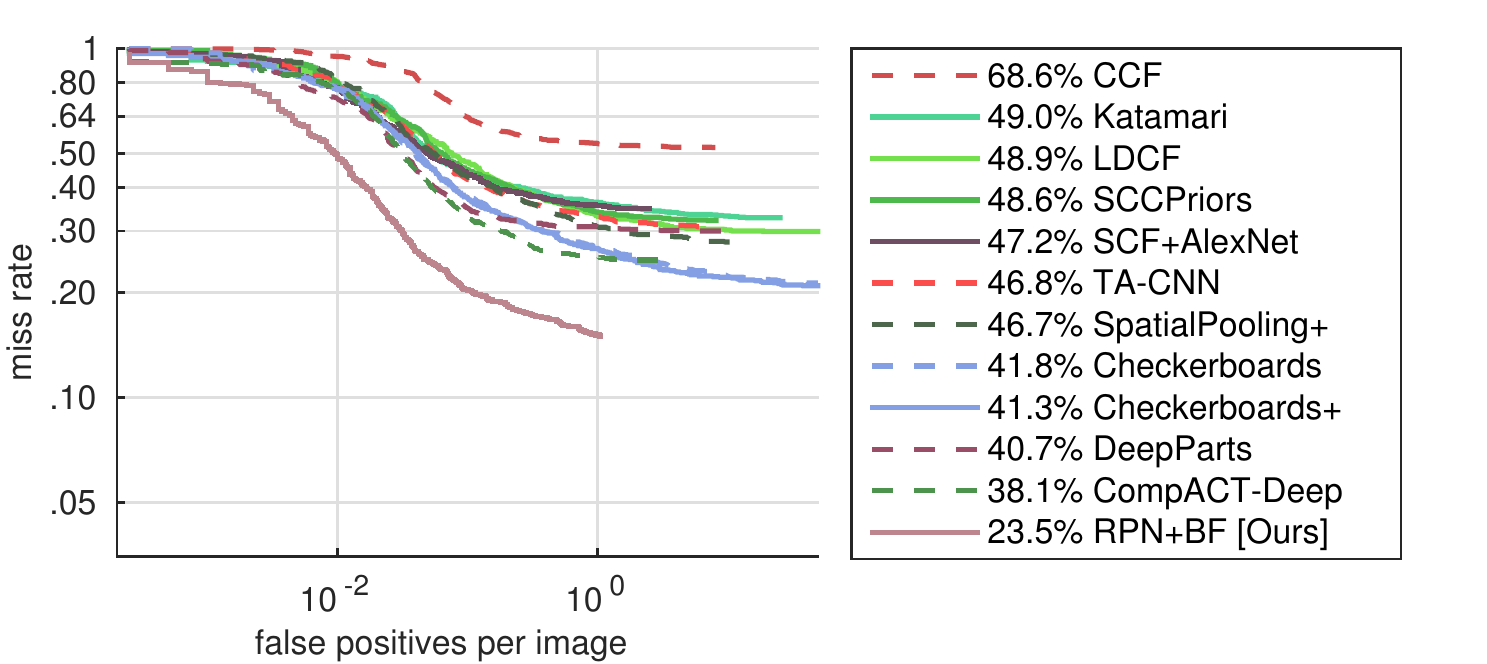}
\caption{Comparisons on the \textbf{Caltech} set using an IoU threshold of 0.7 to determine True Positives (legends indicate MR).}
\label{fig:caltech_iou70}
\end{figure}

To verify that the bootstrapping scheme in BF is of central importance (instead of the tree structure of the BF classifiers), we replace the last-stage BF classifier with a Fast R-CNN classifier. The results are in Table~\ref{table-bootstrap}.
Formally, after the 6 stages of bootstrapping, the bootstrapped training set is used to train a Fast R-CNN classifier (instead of the final BF with 2048 trees). We perform this comparison using RoI features on Conv5\_3 (\`a trous). The bootstrapped Fast R-CNN has an MR of 14.3\%, which is closer to the BF counterpart of 13.7\%, and better than the non-bootstrapped Fast R-CNN's 16.2\%. This comparison indicates that the major improvement of BF over Fast R-CNN is because of bootstrapping, whereas the shapes of classifiers (forest \vs MLP) are less important.

\begin{figure}[t] \centering
\includegraphics[width=1\textwidth]{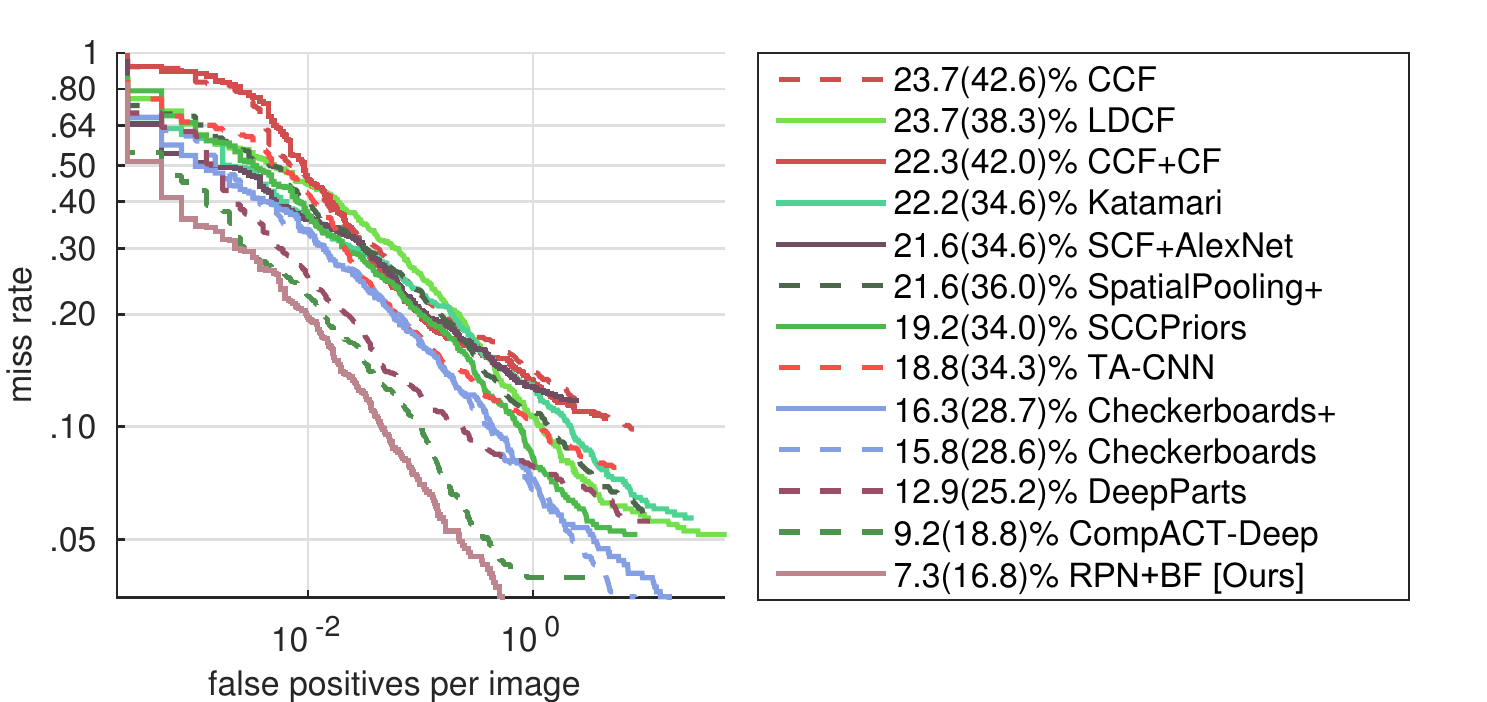}
\caption{Comparisons on the \textbf{Caltech-New} set (legends indicate MR$_{-2}$ (MR$_{-4}$)).}
\label{fig:caltech-new}
\end{figure}

\subsection{Comparisons with State-of-the-art Methods}

\begin{figure}[t] \centering
\includegraphics[width=1\textwidth]{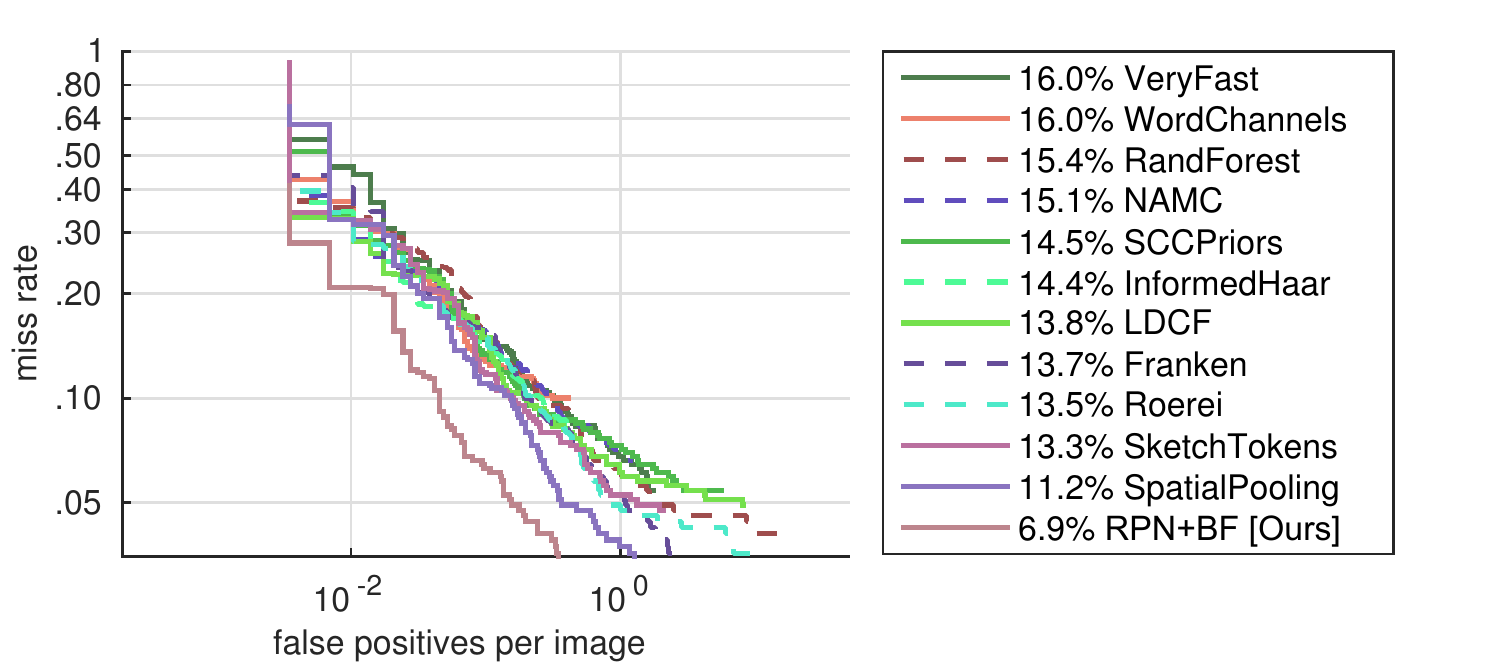}
\caption{Comparisons on the \textbf{INRIA} dataset (legends indicate MR).}
\label{fig:inria}
\includegraphics[width=.95\textwidth]{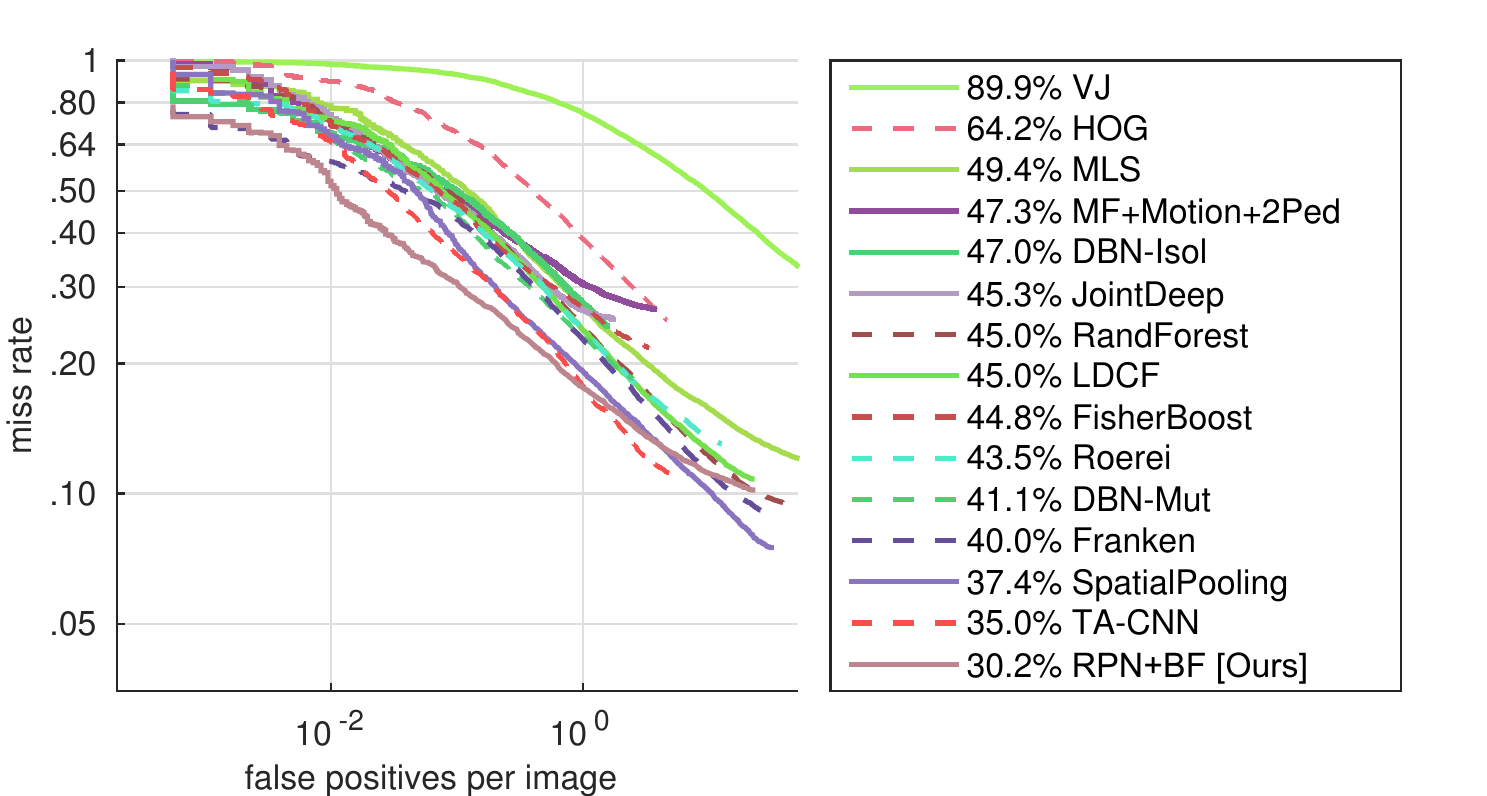}
\caption{Comparisons on the \textbf{ETH} dataset (legends indicate MR).}
\label{fig:eth}
\end{figure}

\subsubsection{Caltech} Fig.~\ref{fig:caltech} and  \ref{fig:caltech-new} show the results on Caltech. In the case of using original annotations (Fig.~\ref{fig:caltech}), our method has an MR of \textbf{9.6\%}, which is over 2 points better than the closest competitor (11.7\% of CompactACT-Deep \cite{cai2015learning}). In the case of using the corrected annotations (Fig.~\ref{fig:caltech-new}), our method has an MR$_{-2}$ of 7.3\% and MR$_{-4}$ of 16.8\%, both being 2 points better than the previous best methods.

In addition, expect for CCF (MR 18.7\%) \cite{yang2015convolutional}, ours (MR 9.6\%) is the only method  that \emph{uses no hand-crafted features}. Our results suggest that hand-crafted features are not essential for good accuracy on the Caltech dataset; rather, high-resolution features and bootstrapping are the key to good accuracy, both of which are missing in the original Fast R-CNN detector.

Fig.~\ref{fig:caltech_iou70} shows the results on Caltech where an IoU threshold of 0.7 is used to determine True Positives (instead of 0.5 by default). With this more challenging metric, most methods exhibit dramatic performance drops, \eg, the MR of CompactACT-Deep \cite{cai2015learning}/DeepParts \cite{tian2015deep} increase from 11.7\%/11.9\% to 38.1\%/40.7\%. Our method has an MR of 23.5\%, which is \textbf{a relative improvement of $\sim$40\%} over the closest competitors. This comparison demonstrates that our method has a substantially better \textbf{localization} accuracy. It also indicates that there is much room to improve localization performance on this widely evaluated dataset.

Table~\ref{table-time} compares the running time on Caltech. Our method is as fast as  CompACT-Deep \cite{cai2015learning}, and is much faster than CCF \cite{yang2015convolutional} that adopts feature pyramids. Our method shares feature between RPN and BF, and achieves a good balance between speed and accuracy.

\renewcommand\arraystretch{1.25}
\setlength{\tabcolsep}{6pt}
\begin{table}[t]
\begin{center}
\begin{tabular}{c|c|c|c}
\hline
method  & hardware & time/img (s) & MR (\%) \\
\hline\hline
LDCF \cite{nam2014local}  & CPU & 0.6 & 24.8\\
CCF \cite{yang2015convolutional}  & Titan Z GPU & 13 & 17.3 \\
CompACT-Deep \cite{cai2015learning}  & Tesla K40 GPU & \textbf{0.5} & 11.7 \\
\hline
RPN+BF [ours]  & Tesla K40 GPU & \textbf{0.5} & \textbf{9.6} \\
\hline
\end{tabular}
\end{center}
\caption{Comparisons of running time on the Caltech set. The time of LDCF and CCF is reported in \cite{yang2015convolutional}, and that of CompactACT-Deep is reported in \cite{cai2015learning}.}
\label{table-time}
\end{table}

\subsubsection{INRIA and ETH} Fig.~\ref{fig:inria} and \ref{fig:eth} show the results on the INRIA and ETH datasets. On the INRIA set, our method achieves an MR of 6.9\%, considerably better than the best available competitor's 11.2\%. On the ETH set, our result (30.2\%) is better than the previous leading method (TA-CNN \cite{tian2015pedestrian}) by 5 points.

\subsubsection{KITTI} Table~\ref{table-kitti} shows the performance comparisons on KITTI. Our method has competitive accuracy and fast speed.

\begin{table}[t]
\begin{center}
\begin{tabular}{c|c|c|c|c}
\hline
method & mAP  on Easy &  mAP on  Moderate & mAP on  Hard & Times (s) \\
\hline\hline
R-CNN & 61.61 & 50.13 & 44.79 & 4 \\
pAUCEnstT & 65.26 & 54.49 & 48.60 & 60 \\
FilteredICF & 67.65 & 56.75 & 51.12 & 2 \\
DeepPart & 70.49 & 58.67 & 52.78 & 1 \\
CompACT-Deep & 70.69 & 58.74 & 52.71 & 1 \\
Regionlets & 73.14 & \textbf{61.15} & \textbf{55.21} & 1$^\dag$ \\
\hline
RPN+BF [ours]  & \textbf{77.12} & \textbf{61.15} & \textbf{55.12} & 0.6 \\
\hline
\end{tabular}
\end{center}
\caption{Comparisons on the \textbf{KITTI} dataset collected at the time of submission (Feb 2016). The timing records are collected from the KITTI leaderboard. $^\dag$: region proposal running time ignored (estimated 2s).}
\label{table-kitti}
\end{table}

\section{Conclusion and Discussion}

In this paper, we present a very simple but effective baseline that uses RPN and BF for pedestrian detection. On top of the RPN proposals and features, the BF classifier is flexible for (i) combining features of arbitrary resolutions from any layers, without being limited by the classifier structure of the pre-trained network; and (ii) incorporating effective bootstrapping for mining hard negatives. These nice properties overcome two limitations of the Faster R-CNN system for pedestrian detection. Our method is a self-contained solution and does not resort to hybrid features.

Interestingly, we show that \emph{bootstrapping} is a key component, even with the advance of deep neural networks. Using the same bootstrapping strategy and the same RoI features, both the tree-structured BF classifier and the region-wise MLP classifier (Fast R-CNN) are able to achieve similar results (Table~\ref{table-bootstrap}).
Concurrent with this work, an independently developed method called Online Hard Example Mining (OHEM) \cite{shrivastava2016training} is developed for training Fast R-CNN for general object detection. It is interesting to investigate this end-to-end, online mining fashion \vs the multi-stage, cascaded bootstrapping one.

\subsubsection{Acknowledgement} This work was supported in part by State Key Development Program under Grant 2016YFB1001000, in part by Guangdong Natural Science Foundation under Grant S2013050014548. This work was also supported by Special Program for Applied Research on Super Computation of the NSFC-Guangdong Joint Fund (the second phase). We thank the anonymous reviewers for their constructive comments on improving this paper. 

\bibliographystyle{unsrt}
\bibliography{pedestrian}
\end{document}